\documentclass[sigconf]{acmart}

\makeatletter
\def\@ACM@checkaffil{
    \if@ACM@instpresent\else
    \ClassWarningNoLine{\@classname}{No institution present for an affiliation}%
    \fi
    \if@ACM@citypresent\else
    \ClassWarningNoLine{\@classname}{No city present for an affiliation}%
    \fi
    \if@ACM@countrypresent\else
        \ClassWarningNoLine{\@classname}{No country present for an affiliation}%
    \fi
}
\makeatother

\usepackage{multirow}
\usepackage{multicol}
\usepackage{graphicx}

\usepackage{amssymb}
\usepackage{bbding}
\usepackage{wasysym}
\usepackage{colortbl}
\usepackage{xcolor}
\usepackage{algorithm}
\usepackage{algorithmic}
\usepackage{setspace}
\makeatletter
\newcommand\tabcaption{\def\@captype{table}\caption}
\newcommand\figcaption{\def\@captype{figure}\caption}
\makeatother
\AtBeginDocument{%
  \providecommand\BibTeX{{%
    \normalfont B\kern-0.5em{\scshape i\kern-0.25em b}\kern-0.8em\TeX}}}

\copyrightyear{2023}
\acmYear{2023}
\setcopyright{acmlicensed}
\acmConference[MM '23]{Proceedings of the 31st ACM International Conference on Multimedia}{October 29-November 3, 2023}{Ottawa, ON, Canada}
\acmBooktitle{Proceedings of the 31st ACM International Conference on Multimedia (MM '23), October 29-November 3, 2023, Ottawa, ON, Canada}
\acmPrice{15.00}
\acmDOI{10.1145/3581783.3611889}
\acmISBN{979-8-4007-0108-5/23/10}

\begin{document}

\title[Improving Scene Graph Generation with Superpixel-Based Interaction Learning]{Improving Scene Graph Generation \\ with Superpixel-Based Interaction Learning}

\author{Jingyi Wang}
\affiliation{%
  \institution{Tsinghua University}}
\email{wang-jy20@mails.tsinghua.edu.cn}

\author{Can Zhang}
\affiliation{{
Tencent Media Lab
}}
\email{cannyzhang@tencent.com}

\author{Jinfa Huang}
\affiliation{{
University of Rochester
}}
\email{jhuang90@ur.rochester.edu}

\author{Botao Ren}
\affiliation{%
  \institution{Tsinghua University}}
\email{rbt22@mails.tsinghua.edu.cn}

\author{Zhidong Deng}
\authornote{Corresponding author. Zhidong Deng is with Beijing National Research Center for Information
Science and Technology (BNRist), Institute for Artificial Intelligence at Tsinghua University (THUAI), Department of Computer Science, State Key Laboratory of Intelligent Technology and Systems, Tsinghua University, Beijing 100084, China.}
\affiliation{%
  \institution{Tsinghua University}}
\email{michael@mail.tsinghua.edu.cn}


\begin{abstract}
 Recent advances in Scene Graph Generation (SGG) typically model the relationships among entities utilizing box-level features from pre-defined detectors. We argue that an overlooked problem in SGG is the coarse-grained interactions between boxes, which inadequately capture contextual semantics for relationship modeling, practically limiting the development of the field. 
In this paper, we take the initiative to explore and propose a generic paradigm termed \underline{S}uperpixel-based \underline{I}nteraction \underline{L}earning (SIL) to remedy coarse-grained interactions at the box level. It allows us to model fine-grained interactions at the superpixel level in SGG. Specifically, (i) we treat a scene as a set of points and cluster them into superpixels representing sub-regions of the scene. (ii) We explore intra-entity and cross-entity interactions among the superpixels to enrich fine-grained interactions between entities at an earlier stage. 
Extensive experiments on two challenging benchmarks (Visual Genome and Open Image V6) prove that our SIL enables fine-grained interaction at the superpixel level above previous box-level methods, and significantly outperforms previous state-of-the-art methods across all metrics. 
More encouragingly, the proposed method can be applied to boost the performance of existing box-level approaches in a plug-and-play fashion. In particular, SIL brings an average improvement of 2.0\% mR (even up to 3.4\%) of baselines for the PredCls task on Visual Genome, which facilitates its integration into any existing box-level method.
\end{abstract}

\ccsdesc[500]{Computing methodologies~Scene understanding}

\keywords{scene graph, superpixel, clustering, visual understanding}


\maketitle

\section{Introduction}
\vspace{-0.2cm}
\begin{figure}[h]
  \centering
    \includegraphics[width=1.0\linewidth]{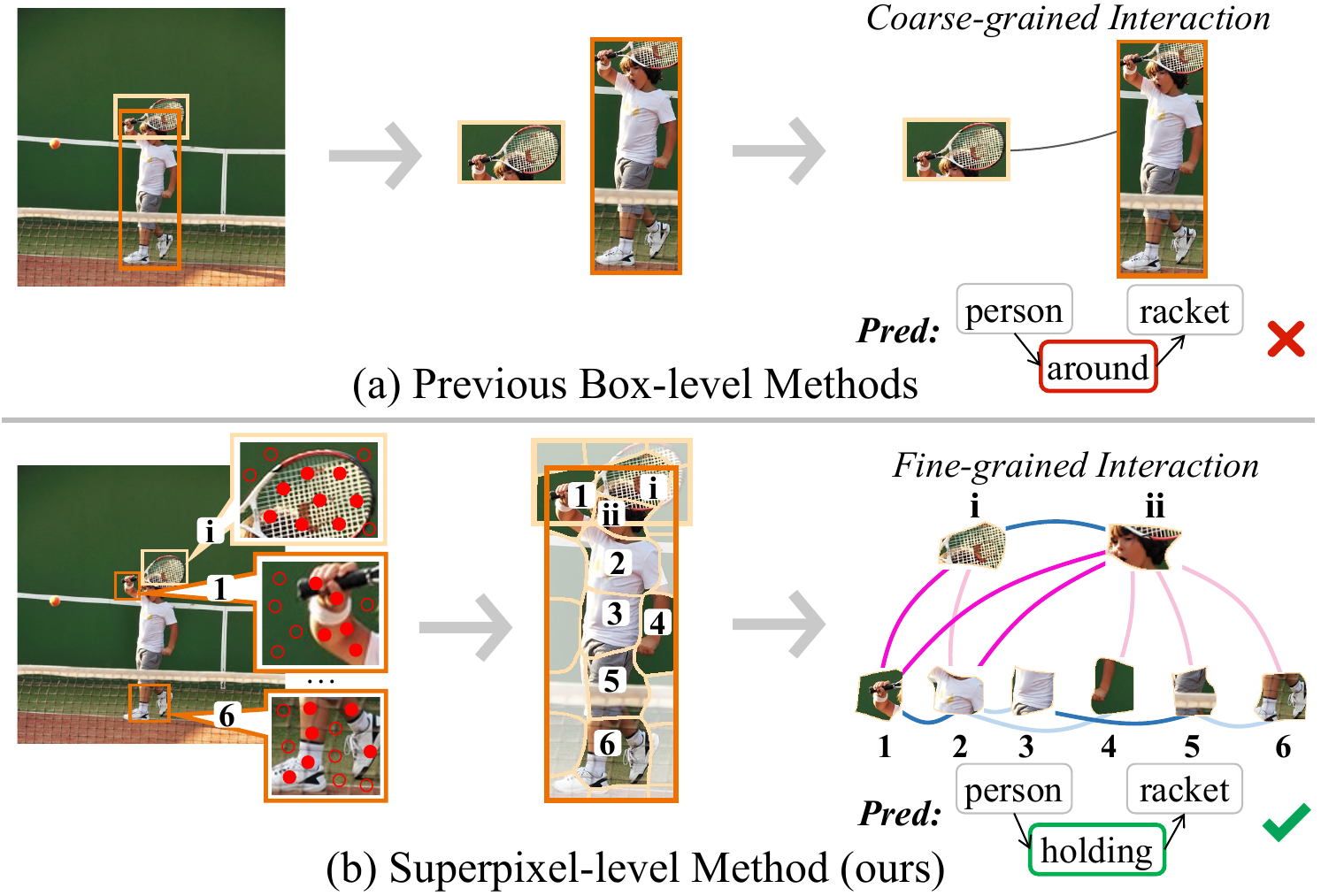}
  \caption{The illustration of the difference between (a) previous box-level methods and (b) our superpixel-level method. Previous methods only consider coarse-grained box-level interactions. In our method, we treat the image as a set of superpixels and extract visual features using the clustering algorithm. We explore the intra-entity interaction and cross-entity interaction at the superpixel level, obtaining fine-grained interaction information.}
  \label{fig-motivation}
\end{figure}
\vspace{-0.3cm}
Scene graphs depict multiple entities and their interaction relationships in scenes with a graph data format. 
In a scene graph, entities in the original scene are represented as nodes, and relationships among these nodes are represented as edges among nodes. 
These entities and relationships form <subject-predicate-object> triplets in the scene graph. 
The understanding and reasoning capabilities of scenes conveyed in the scene graph have driven researchers to further explore the generation and application of scene graphs.
Scene graphs are useful for performing various tasks such as visual question answering\cite{vqa2}, image captioning\cite{caption1,caption2}, and other application scenarios\cite{survey,mm6,mm7}, all of which are helpful for visual understanding\cite{hhh1,hhh2,zzz1,zzz2}.

The typical pipeline of SGG includes three steps. The first step is using an object detector to identify entities in the image, getting the bounding boxes and classification labels of entities. Then, the box-level features from the detector output are adopted to model the interactions between entities. Finally, we classify the relationships between entities and construct the scene graph with detected entity nodes and classified relations.
Existing SGG methods have focused on modeling interactions and utilizing contextual information, which is crucial for generating scene graphs\cite{sg4,sg5,squat}. However, the existing methods rely on box-level features and interactions as shown in Fig. \ref{fig-motivation}(a). Such coarse-grained interactions result in two drawbacks.
Firstly, at the entity level, existing box-level methods can not model the interactions between different sub-regions inside entities. The box-level features they used are just rough representations of the rectangular box, without distinguishing different sub-regions within the bounding box explicitly. Such features can not capture the important sub-region within the entities.
Secondly, at the relation level, researchers have overlooked more fine-grained interactions among entities. With the rough box-level features, the later relation prediction modules inadequately capture contextual semantics for relationship modeling. Besides, such coarse-grained interactions among boxes would be confused by redundant background information, failing to focus on the specific location where the relation between entities occurs.

To address the aforementioned limitations in SGG brought by box-level features, we reformulate the existing pipeline into a new superpixel-level SGG process as shown in Fig. \ref{fig-motivation}(b). This process aims to treat an entity as a set of sub-regions. In our method, an image is viewed as a set of points first and the points are clustered into superpixels as the expression of sub-regions.
With regard to the rough box-level features, superpixels refine entity representation, allowing for a more fine-grained analysis of the interactions between entities. 
In this way, we obtain more specific entity features for the SGG task. At the relation level, we inject intra-entity and cross-entity interaction into superpixels. In Fig. \ref{fig-motivation}(b), pink curves indicate cross-entity interactions among superpixels, and blue curves indicate intra-entity interactions. Light-colored curves indicate weak interactions. As we know, we are the first to view entities as sets of superpixels and modify the previous box-based methods with coarse-grained interactions into superpixel-based methods with fine-grained interactions in the field of scene graph generation.

Accordingly, we propose a Superpixel-based Interaction Learning (SIL) block, converting box-level features into superpixel-level features and injecting intra-entity and cross-entity interactions into superpixels for the SGG task. 
First, SIL views an entity as a set of sub-regions and uses superpixels to express them. We aggregate the pixels in an image or the elements in a feature map using clustering algorithm. The cluster results form multiple superpixels as the expression of the image. Equipped with SIL, existing methods are strengthened with more fine-grained visual features.
Second, with a box as multiple irregular sub-regions, we model intra-entity interactions, aggregating the superpixels inside an entity. We assign different weights to different superpixels to emphasize their varying importance for the corresponding entity and calibrate superpixel features.
Third, as a remedy for coarse-grained interactions at the box level in previous methods, we inject cross-entity interactions into superpixels. This step allows the model to benefit from the fine-grained interactions between superpixels and help the judgment of entity-level relations by focusing on specific sub-regions related to predicates. Besides, this step enables us to start modeling interactions between entities at an earlier stage in the SGG pipeline.
Finally, the enhanced visual features are fed into an arbitrary relation prediction module of previous SGG methods to construct scene graphs.

We evaluate our SIL on the Visual Genome (VG) \cite{vg} and Open Images V6 (OI V6) \cite{oi} datasets using typical and latest SGG baselines. Extensive experiments demonstrate the effectiveness and generalization ability of our SIL method. On VG dataset, the SIL block brings an average improvement of 2.0\% to the mean Recall of baselines on the PredCls task. With SIL, we outperform previous state-of-the-art methods on VG and OI V6 datasets. Our main contributions are summarized as follows: 
(1) To the best of our knowledge, we are the first to reformulate the existing box-based SGG pipeline into a new superpixel-level SGG process. We regard a scene as a set of superpixels obtained using the clustering algorithm and adapt the features of these superpixels to the final scene graph generation task. 
(2) We model intra-entity and cross-entity interactions between the superpixels to perform fine-grained relation modeling among entities, helping the judgment of relationships between entities of scene graphs. 
(3) We propose a plug-and-play Superpixel-based Interaction Learning block, which improves the performance of the mainstream SGG methods stably on the public VG and OI V6 datasets.

\section{Related Work}
\begin{figure*}[htbp]
    \centering
    \includegraphics[width=\textwidth]{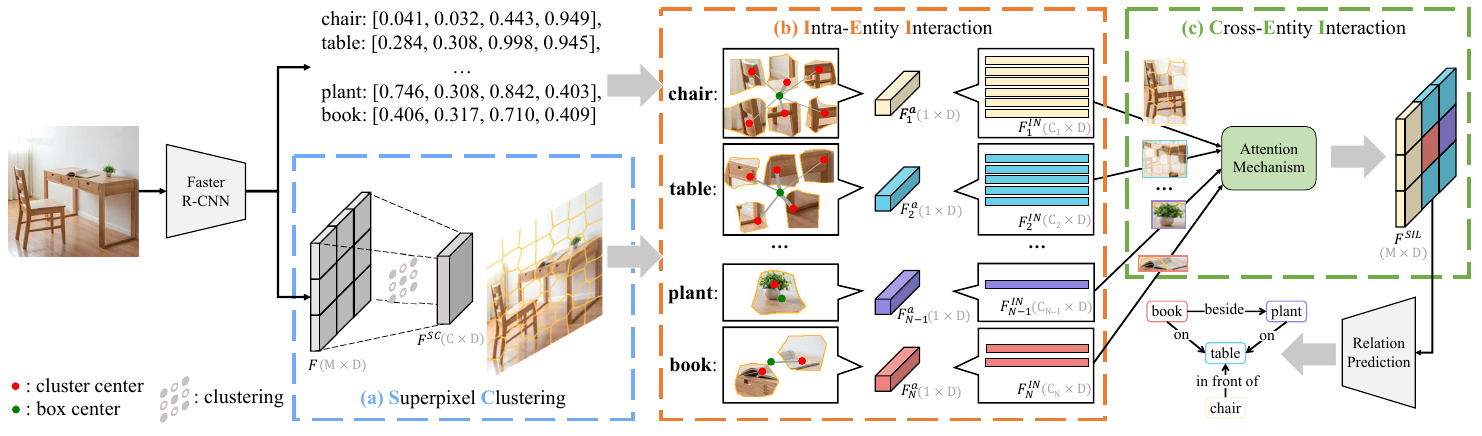}
    \caption{The structure of our Superpixel-based Interaction Learning (SIL) block and its application in Scene Graph Generation (SGG). In SGG, we first detect the entities in the input image, obtaining their bounding boxes and the feature map of the image. With the proposed SIL block, we perform clustering on pixels of the image to obtain superpixel-level features as depicted in (a). With the detected boxes and superpixel-level features, we perform Intra-Entity Interaction in (b) and Cross-Entity Interaction in (c). Finally, we perform relation prediction as usual to construct scene graphs.}
    \label{fig-block}
\end{figure*}
\subsection{Scene Graph Generation}
A scene graph is a graphical representation that captures the entities and their relationships in a scene, with nodes representing the entities and edges denoting the relationships between them. Scene graph generation (SGG) has extracted significant research interests due to its potential to facilitate visual comprehension of images and videos\cite{mm2, mm3,cvpr1,cvpr4, rebuttal2}, and help scene understanding\cite{rebuttal7,rebuttal8}. 
There are also methods for video and using of distillation\cite{cvpr5,r3_1,r3_2,r3_3,r3_4}. 
The typical SGG pipeline includes three stages. First, detect entities in a scene and extract visual features of entities. Second, associate entities and enhance the contextual information between them. Third, classify the relations between entities and construct a scene graph with detected entities and classified relations.

Numerous methods have been proposed to improve SGG. Most of them focus on the last two stages mentioned above, namely contextualization and relation classification\cite{zzz3,bgnn,mm4,mm5,sg9}. 
As for the contextualization stage, \cite{motif} adopts the object context to compute and propagated contextual information among objects, and constructs edge context that gives global context for edge predictions. 
Researchers fuse the feature representations of entities and relations for SGG\cite{sgfuse1}. Cross-modal methods are also been applied and borrowed\cite{icra,hhh3, hhh4}.
As for the relation classification and graph construction stage, researchers try to tackle the long-tail and biasing problem\cite{mm1, mm4, cvpr2, cvpr3}. \cite{pcpl} adjust loss weights by perceiving and utilizing the correlation among all predicate classes. 
\cite{notall} exploits relation labels based on their informativeness. The authors of \cite{stacked} use a group of classifiers that are expert in distinguishing different subsets of classes. 

As we can see, the majority of the current methods concentrate on the contextualization and relation classification stage. 
Existing methods simply use the box-level features without finer interaction modeling whether they are bottom-up two-stage or end-to-end structures. Such coarse-grained interactions between boxes inadequately capture contextual semantics and limit the development of the SGG field.
They ignore the role of interactions between sub-regions within boxes in scene graph generation. Addressing this deficiency, we reformulate the existing pipeline into a new superpixel-level SGG process and explore fine-grained interactions between entities.

\subsection{Superpixel Algorithm}
SuperPixel groups pixels with similar features in an image to form a more representative pixel using the clustering algorithm\cite{superpixel}. The grouped superpixel can be used as the basic unit for other image processing algorithms. 
To reduce the computational costs of the original SuperPixel method, the Simple Linear Iterative Clustering (SLIC) algorithm performs local clustering of image pixels, resulting in compact and approximately uniformly-sized superpixels\cite{slic}.
There are also some superpixel segmentation methods based on other structures such as linear spectral clustering\cite{sp2}, convolutional networks\cite{sp4,sp5}, and spanning forest\cite{sp6}.
Recently, researchers propose using the clustering algorithm to group and extract deep features of images hierarchically\cite{coc}. It relies on the clustering algorithm for spatial interaction and provides a new perspective to interpret images. CLUSTSEG\cite{add1} proposes a transformer-based framework that tackles different image segmentation tasks\cite{r3_5segmentation} including superpixel through a clustering scheme.

\section{Method}
In this section, we first introduce the basic workflow of the SGG task and the drawbacks of previous SGG methods. 
Then, we introduce the plug-and-play Superpixel-based Interaction Learning (SIL) block for scene graph generation. The overview of our SIL is shown in Fig. \ref{fig-block} and Algorithm \ref{alg}.
The SIL block takes an image or its feature map as the input. 
The clustering step in Fig. \ref{fig-block}(a) performs clustering on the pixels or elements of the input to split it into multiple sub-regions based on their visual features. The orange splits in the image of Fig. \ref{fig-block}(a) show the clustering result. 
Then, we model the interactions inside each entity as shown in Fig. \ref{fig-block}(b). Superpixels inside a box are aggregated with different weights to emphasize their varying importance for the corresponding entity.
In Fig. \ref{fig-block}(b), green circles indicate the centers of each bounding box, and red circles indicate the centers of each cluster. For each entity, we calculate the distance between the center of the box and each cluster center within the entity. Clusters at far distances have low weights.
The aggregated feature are then dispatched to superpixels to enhance interactions within each entity. Tensors with different colors in Fig. \ref{fig-block}(b) indicate features of different entities.
After that, we inject interactions into all superpixels to achieve fine-grained cross-entity interactions to serve the SGG task as shown in Fig. \ref{fig-block}(c). Interactions among superpixels of different entities imply the interaction between entities.
These three modules in the SIL block will be clarified in the latter subsections.

\begin{algorithm}[htbp]\small
\small
\caption{\small{The proposed superpixel-based interaction learning block.}}
\label{alg}
\begin{algorithmic}[1]
\REQUIRE  Image features $F \in \mathbb{R}^{M \times D} $ ,\\
Entity boxes $b=b_1, b_2, \cdots, b_N$,\\
Entity reference features $g = g_1, g_2, \cdots, g_N$,\\
Points $Q \in \mathbb{R}^{M \times 2}$, Centers $T \in \mathbb{R}^{C \times 2}$ \\
\textbf{\# Superpixel Clustering}
\FOR{each center $T_j$ in $T$}
\STATE $F^{c}_{j} \gets \sum_{Q_i \in {\rm{KNN}} (T_j)} F_{i}/n_k \qquad \qquad \qquad \qquad \qquad \quad \; \; \;$  \# Eq.~\eqref{equ-fctr}
\ENDFOR
\FOR{each point $Q_i$ in $Q$}
\STATE  $P^{QT}_i  \gets {\rm argmax}_j sim(F_i, F^{c}_j)\qquad \qquad \qquad \qquad \qquad \; \; \; $\# Eq.~\eqref{equ-pqt}
\ENDFOR
\FOR{each center $T_j$ in $T$}
\STATE $F^{SC}_j = H^{\prime}(F^{c}_j, P^{QT}, F)\qquad \qquad \qquad \qquad \qquad \qquad \quad$ \# Eq.~\eqref{fctra}
\ENDFOR

\textbf{\# Intra-Entity Interaction}
\FOR{each entity box $b_k$ in $b$}
\STATE $P^{BT}_k \gets \{
    T_j: (x^T_j , y^T_j) {\rm \; falls \; within \;} b_k \} $
\STATE $F^{a}_k = H^{a}(g_k, P^{BT}_k,b_k,T,F^{SC})\qquad \qquad \qquad \qquad \qquad $  \# Eq.~\eqref{equ-fagg}
\STATE $F^{IN}_k \gets \emptyset$ 
\FOR{each center $T_j$ in $P^{BT}_k$}
\STATE $F^{IN}_k \gets F^{IN}_k \cup \{F_{j}^{SC} + F_{k}^{a}  \times w(j,k)\}\qquad \qquad \qquad $\# Eq.~\eqref{equ-fr}
\ENDFOR
\ENDFOR

\textbf{\# Cross-Entity Interaction}
\STATE $F^{CR} \gets {\rm Multihead}(F^{IN}, F^{IN},F^{IN})\qquad \qquad \qquad \qquad \quad   $ \# Eq.~\eqref{equ-fa}
\FOR {each point $Q_i$ in $Q$}
\STATE $F^{SIL}_i = H^{\prime \prime}(F_i, P^{QT}_i,F^c,F^{CR}) \qquad \qquad \qquad \qquad \qquad$ \# Eq.~\eqref{equ-fsil}
\ENDFOR
\RETURN  $F^{SIL}$
\end{algorithmic}
\end{algorithm}
\vspace{-0.5cm}

\subsection{Problem Definition}
Scene graph generation takes an image as the input and generates a graph $G = (O, R)$, where $O$ is the set of entities as nodes of the scene graph and $R$ is the set of relations as edges among nodes in $O$. The object detector provides the feature map $F$ of the input image. Suppose $O= \{ o_1, o_2, \cdots, o_N  \}$ with $N$ as the number of entities in the input image. 
We use $b=\{b_1, b_2, \cdots, b_N\}$ to denote the bounding boxes of entites in $O$. Entities in $O$ and relationships in $R$ form multiple <subject-predicate-object> triplets and construct a scene graph of the input image. 
Existing methods align the box coordinates in $b$ with the feature map $F$ to obtain the box-level feature $f_k$ for each $o_k$. $f_k$ of entities are usually adopted directly as the input of the relation prediction module. However, $f_k$ lacks differentiation of sub-regions within $b_k$ and disables more fine-grained interaction between sub-regions within boxes.

Therefore, we propose a Superpixel-based Interaction Learning framework for SGG. In the following subsections, we will introduce the overall structure of our SIL and clarify its key modules.

\subsection{Superpixel Clustering}
\label{sec-3.1}
Unlike previous methods based on box-level features, SIL considers an image as a set of points and clusters these points to form superpixels. Correspondingly, in Fig. \ref{fig-block}(a), we convert the format of the input as follows. First, with $M=w\times h$, we view pixels in the input feature map as $F \in \mathbb{R}^{M \times D}$ with $M$ points. We use $Q = \{ ( x^Q_1, y^Q_1 ), ( x^Q_2, y^Q_2 ), \cdots, ( x^Q_M, y^Q_M )  \}$ to denote the 2-D coordinates of these $M$ points. Second, we propose $C$ centers denoted by $T = \{ ( x^T_1, y^T_1 ), ( x^T_2, y^T_2 ), \cdots, ( x^T_C, y^T_C )  \}$ as the clustering centers of the points in $Q$. As usual, we propose centers on the two dimensions of the 2D input evenly and separately. Then we pair up the centers of the two dimensions to obtain all the final centers. For example, if 7 clustering centers are assigned for each dimension, 49 centers will be found eventually.
Following \cite{slic}, we average the features of the nearest points with each center in $T$ to obtain the feature of centers denoted by $F^{c} \in \mathbb{R}^{C \times D}$, which is calculated with:
\begin{equation}
\label{equ-fctr}
    F^{c}_{j} = \frac{1}{n_k} \sum_{Q_i \in {\rm{KNN}} (T_j)} F_{i},
\end{equation}
where ${\rm{KNN}}(T_j)$ denotes the $n_k$-Nearest Neighbors of $T_j$  and $n_k$ is the number of the nearest neighbours we use. Third, we perform clustering on these points. Specifically, we evaluate the pairwise similarity between the point features $F$ and the center features $F^{c}$. According to the similarities, the $M$ points in $Q$ are grouped into the nearest clusters in $T$. We denote the clustered results from $Q$ to $T$ as $P^{QT} = \{P^{QT}_1, P^{QT}_2, \cdots, P^{QT}_M\}$ and we can get:
\begin{equation}
\label{equ-pqt}
     P^{QT}_i = { \underset {j\in [1,C]} { \operatorname {argmax} }} \; sim(F_i, F^{c}_j),
\end{equation}
where $sim$ calculates the cosine similarity. The value of $P^{QT}_i$ indicates which cluster point $Q_i$ belongs to. In this way, the input is converted into $C$ clusters and each cluster corresponds to a superpixel in the image. 

After the clustering step, we update the features of clustering centers in $T$. To be specific, we aggregate features of points in a cluster and update the feature of the cluster center, that is, we update $F^{c}$ with $F$ in each cluster to get $F^{SC} \in \mathbb{R}^{C \times D}$. The calculation of center $T_j$ of $F^{SC}$ can be formulated as:
\begin{equation}
\label{fctra}
    F^{SC}_j = H^{\prime} \Big( \frac{
F^{c}_j+ \sum_{i} ( F_i \times sim(F_i, F^{c}_j) \times  {\rm \textbf{1}}(P^{QT}_i=j))    }{
      1  + \sum_{i} (sim(F_i, F^{c}_{j})\times {\rm \textbf{1}}(P^{QT}_i=j)))
    }  \Big),
\end{equation}
where $H^{\prime}$ is a network for feature dimension transformation and $\rm \textbf{1}(\cdot)$ is an indicator function. 
After clustering and superpixel-level feature extracting, $F^{SC}$ provides finer superpixel-level visual information than $F$. The updated center features $F^{SC}$ would participate in the Intra-Entity Interaction module as shown in Fig. \ref{fig-block}(b). 

\begin{table*}[htbp]
\centering
\caption{Mean Recall (\%) comparison results on the VG dataset over three tasks.  $^{\ast}$ denotes the results reproduced using the official trained model or released code.}
\label{tab-exp}
\begin{tabular}{c|lll|lll|lll}
\hline
\multicolumn{1}{c|}{\multirow{2}{*}{Model}} & \multicolumn{3}{c|}{PredCls} & \multicolumn{3}{c|}{SGCls} & \multicolumn{3}{c}{SGGen} \\
\multicolumn{1}{c|}{}& \multicolumn{1}{c}{mR@20}   & \multicolumn{1}{c}{mR@50}   & \multicolumn{1}{c|}{mR@100}   & \multicolumn{1}{c}{mR@20}   & \multicolumn{1}{c}{mR@50}  & \multicolumn{1}{c|}{mR@100}  & \multicolumn{1}{c}{mR@20}  & \multicolumn{1}{c}{mR@50}  & \multicolumn{1}{c}{mR@100}  \\ \hline
MOTIFS\cite{motif}\tiny{\emph{CVPR'18}}    & 11.7& 14.8& 16.1& 6.3 & 8.3& 8.8& 5.0& 6.8& 7.9\\
\rowcolor{gray!10} MOTIFS+SIL         &\textbf{13.6} {\footnotesize \textbf{\textcolor{ACMRed}{(+1.9)}}}& \textbf{16.9} {\footnotesize \textbf{\textcolor{ACMRed}{(+2.1)}}}& \textbf{18.4} {\footnotesize \textbf{\textcolor{ACMRed}{(+2.3)}}}& \textbf{8.5} {\footnotesize \textbf{\textcolor{ACMRed}{(+2.2)}}}&\textbf{10.2} {\footnotesize \textbf{\textcolor{ACMRed}{(+1.9)}}}& \textbf{10.8} {\footnotesize \textbf{\textcolor{ACMRed}{(+2.0)}}}& \textbf{5.5} {\footnotesize \textbf{\textcolor{ACMRed}{(+0.5)}}}&\textbf{7.3} {\footnotesize \textbf{\textcolor{ACMRed}{(+0.5)}}}& \textbf{8.5} {\footnotesize \textbf{\textcolor{ACMRed}{(+0.6)}}}   \\ \cline{1-10} 

$^{\ast}$G-RCNN\cite{graphrcnn}\tiny{\emph{ECCV'18}} & 13.5 & 16.7 & 17.4 & 7.5 & 9.1 & 9.6 & 4.8 & 6.1 & 6.7 \\
\rowcolor{gray!10} G-RCNN+SIL        & \textbf{15.3} {\footnotesize \textbf{\textcolor{ACMRed}{(+1.8)}}} & \textbf{18.1} {\footnotesize \textbf{\textcolor{ACMRed}{(+1.4)}}}& \textbf{19.4} {\footnotesize \textbf{\textcolor{ACMRed}{(+2.0)}}}& \textbf{8.8} {\footnotesize \textbf{\textcolor{ACMRed}{(+1.3)}}}& \textbf{10.7} {\footnotesize \textbf{\textcolor{ACMRed}{(+1.6)}}}& \textbf{11.1} {\footnotesize \textbf{\textcolor{ACMRed}{(+1.5)}}} & \textbf{5.9}  {\footnotesize \textbf{\textcolor{ACMRed}{(+1.1)}}}& \textbf{6.9} {\footnotesize \textbf{\textcolor{ACMRed}{(+0.8)}}}& \textbf{7.6} {\footnotesize \textbf{\textcolor{ACMRed}{(+0.9)}}} \\ \cline{1-10} 

KERN\cite{kern}\tiny{\emph{CVPR'19}}    & 14.7& 17.0& 19.4& 7.4& 9.0& 10.0& 4.8& 6.3& 7.3\\
\rowcolor{gray!10} KERN+SIL           &\textbf{16.0} {\footnotesize \textbf{\textcolor{ACMRed}{(+1.3)}}}& \textbf{18.8} {\footnotesize \textbf{\textcolor{ACMRed}{(+1.8)}}}& \textbf{21.0} {\footnotesize \textbf{\textcolor{ACMRed}{(+1.6)}}}&\textbf{8.4} {\footnotesize \textbf{\textcolor{ACMRed}{(+1.0)}}}& \textbf{10.5} {\footnotesize \textbf{\textcolor{ACMRed}{(+1.5)}}}&\textbf{11.7} {\footnotesize \textbf{\textcolor{ACMRed}{(+1.7)}}}& \textbf{5.3} {\footnotesize \textbf{\textcolor{ACMRed}{(+0.5)}}}& \textbf{6.9} {\footnotesize \textbf{\textcolor{ACMRed}{(+0.6)}}}& \textbf{7.9} {\footnotesize \textbf{\textcolor{ACMRed}{(+0.6)}}}\\ \cline{1-10} 

VCTree\cite{vctree}\tiny{\emph{CVPR'19}}      & 13.2& 16.7& 18.2& 9.6& 11.8& 12.5& 5.4& 7.4& 8.7\\
\rowcolor{gray!10} VCTree+SIL         & \textbf{14.5} {\footnotesize \textbf{\textcolor{ACMRed}{(+1.3)}}}& \textbf{18.1} {\footnotesize \textbf{\textcolor{ACMRed}{(+1.4)}}}& \textbf{19.6} {\footnotesize \textbf{\textcolor{ACMRed}{(+1.4)}}}& \textbf{10.1} {\footnotesize \textbf{\textcolor{ACMRed}{(+0.5)}}}& \textbf{12.6} {\footnotesize \textbf{\textcolor{ACMRed}{(+0.8)}}}& \textbf{13.9} {\footnotesize \textbf{\textcolor{ACMRed}{(+1.4)}}}& \textbf{5.7} {\footnotesize \textbf{\textcolor{ACMRed}{(+0.3)}}}& \textbf{7.6} {\footnotesize \textbf{\textcolor{ACMRed}{(+0.2)}}}& \textbf{8.9} {\footnotesize \textbf{\textcolor{ACMRed}{(+0.2)}}}\\ \cline{1-10} 

GPS-Net\cite{gpsnet}\tiny{\emph{CVPR'20}} & 17.4 &21.3 &22.8 &10.0 &11.8 &12.6 &\textbf{6.9} &8.7 &9.8 \\
\rowcolor{gray!10} GPS-Net+SIL & \textbf{19.1} {\footnotesize \textbf{\textcolor{ACMRed}{(+1.7)}}} & \textbf{23.8} {\footnotesize \textbf{\textcolor{ACMRed}{(+2.5)}}} & \textbf{25.2} {\footnotesize \textbf{\textcolor{ACMRed}{(+2.4)}}} & \textbf{11.7} {\footnotesize \textbf{\textcolor{ACMRed}{(+1.7)}}} & \textbf{13.2} {\footnotesize \textbf{\textcolor{ACMRed}{(+1.4)}}} & \textbf{14.0} {\footnotesize \textbf{\textcolor{ACMRed}{(+1.4)}}} & \textbf{6.9}  {\footnotesize \textbf{\textcolor{ACMRed}{(+0.0)}}} & \textbf{9.1} {\footnotesize \textbf{\textcolor{ACMRed}{(+0.4)}}} & \textbf{10.4} {\footnotesize \textbf{\textcolor{ACMRed}{(+0.6)}}} \\ \cline{1-10}

$^{\ast}$BGNN\cite{bgnn}\tiny{\emph{CVPR'21}} & 24.9  & 29.5 & 31.8 & 12.2 & 14.3 & 14.8 &7.5 &10.4 &12.5 \\
\rowcolor{gray!10} BGNN+SIL           & \textbf{27.9} {\footnotesize \textbf{\textcolor{ACMRed}{(+3.0)}}} & \textbf{32.9} {\footnotesize \textbf{\textcolor{ACMRed}{(+3.4)}}}& \textbf{35.0} {\footnotesize \textbf{\textcolor{ACMRed}{(+3.2)}}}& \textbf{13.1} {\footnotesize \textbf{\textcolor{ACMRed}{(+0.9)}}}& \textbf{15.7} {\footnotesize \textbf{\textcolor{ACMRed}{(+1.4)}}}& \textbf{16.8} {\footnotesize \textbf{\textcolor{ACMRed}{(+2.0)}}}& \textbf{8.2} {\footnotesize \textbf{\textcolor{ACMRed}{(+0.7)}}}& \textbf{11.1} {\footnotesize \textbf{\textcolor{ACMRed}{(+0.7)}}}& \textbf{13.1} {\footnotesize \textbf{\textcolor{ACMRed}{(+0.6)}}}\\ \cline{1-10}

PE-Net\cite{penet}\tiny{\emph{CVPR'23}}  & 25.8 & 31.4 & 33.5 & 15.2   & 18.2   & 19.3    & 9.2    & 12.3   & 14.3    \\
\rowcolor{gray!10} PE-Net+SIL & \textbf{26.9} {\footnotesize \textbf{\textcolor{ACMRed}{(+1.1)}}}    & \textbf{33.1} {\footnotesize \textbf{\textcolor{ACMRed}{(+1.7)}}}    &\textbf{35.3} {\footnotesize \textbf{\textcolor{ACMRed} {(+1.8)}}}    & \textbf{16.7} {\footnotesize \textbf{\textcolor{ACMRed}{(+1.5)}}}   & \textbf{19.9} {\footnotesize \textbf{\textcolor{ACMRed}{(+1.7)}}}   & \textbf{20.7} {\footnotesize \textbf{\textcolor{ACMRed}{(+1.4)}}}  & \textbf{10.0} {\footnotesize \textbf{\textcolor{ACMRed}{(+0.8)}}}   & \textbf{13.0} {\footnotesize \textbf{\textcolor{ACMRed}{(+0.7)}}} & \textbf{15.1} {\footnotesize \textbf{\textcolor{ACMRed}{(+0.8)}}}   \\ \hline
\end{tabular}
\end{table*}

\subsection{Intra-Entity Interaction}
In the Superpixel Clustering step described above, we cluster the image pixels into multiple superpixels as sub-regions and obtain the features $F^{SC}$ as the features of these superpixels. 
However, the feature map is manipulated using only clustering algorithms without any involvement of bounding boxes. Besides, the contribution of each superpixel to the associated entity is not differentiated. Therefore, it is necessary to perform Intra-Entity Interaction as shown in Fig. \ref{fig-block}(b) to emphasize the varying importance of different superpixels and capture finer internal structures and relationships within each entity.
To be specific, we conduct the following operations in the intra-entity interaction module. First, with the detected bounding boxes as the reference, we filter the centers involved in each entity. We denote the results of this filtering process as $P^{BT} = \{P^{BT}_1, P^{BT}_2, \cdots, P^{BT}_{N}\}$. $P^{BT}_k$ consists of all the centers in $T$ where the 2-D coordinates of $T_j$ fall inside $b_k$ in the image. Second, we aggregate the superpixels inside each entity getting $F^{a}\in \mathbb{R}^{N \times D}$ in Fig. \ref{fig-block}(b), which is calculated with:
\begin{equation}
    F^{a}_k = H^{a} \Big(   \frac{
    g_k + \sum_{j} F^{SC}_j \times w(j,k) \times {\rm \textbf{1}}(T_j\in P^{BT}_k)} {
     1 + \sum_{j} d(T_{j}, b_k) \times {\rm \textbf{1}}(T_j\in P^{BT}_k)} \Big),
\label{equ-fagg}
\end{equation}
where $d(T_{j },b_k)$ indicates the Euclidean distance between the 2-D coordinates of $T_{j}$ and the center coordinates of $b_k$. $\rm \textbf{1}(\cdot)$ is an indicator function. $g_k$ in Eq.~\eqref{equ-fagg} is the reference visual feature of $b_k$. $H^{a}$ denotes a network for feature dimension transformation. $w(j,k)$ evaluates the importance of superpixel $T_j$ in entity $o_k$, which is calculated with:
\begin{equation}
\label{equ-weight}
    w(j,k) = \frac{d(T_j, b_k)}{\sum_{j} d(T_{j}, b_k) \times {\rm \textbf{1}}(T_j\in P^{BT}_k)}.
\end{equation}
After that, we use $F^{a}$ to update superpixels inside each entity getting $F^{IN} \in \mathbb{R}^{C \times D}$ in Fig. \ref{fig-block}(b). We have $F^{IN} = \{F^{IN}_1, F^{IN}_2, \cdots, F^{IN}_C\} $ and $F^{IN}_k \in \mathbb{R}^{C_k \times D}$ where $C_k = |P^{BT}_k|$ indicates the number of cluster centers that fall within box $b_k$. The calculation of an entity for $F^{IN}$ can be formulated as:
\begin{equation}
\label{equ-fr}
    F^{IN}_k = \{F_{j}^{SC} + F_{k}^{a}  \times w(j,k)\},
\end{equation}
for all $j$ that satisfy $T_j\in P^{BT}_k$. In this way, we differentiate superpixels in an entity by weights and achieve intra-entity interaction.

\subsection{Cross-Entity Interaction}
Following the above modules in Fig. \ref{fig-block}(a) and (b), the traditional box-level features are transformed into superpixel-level features for SGG through the clustering algorithm and interaction modeling inside each entity. After that, we obtain superpixel features $F^{IN}$ that imply intra-entity interaction. In Fig. \ref{fig-block}(c), we design to inject cross-entity interaction information into superpixels.

Existing methods focus on the interactions at the entity level for SGG. As a remedy for the deficiency of more fine-grained interactions, we inject cross-entity interaction information using $F^{IN}$. In Fig. \ref{fig-block}(c), we collect all the clustering centers in $T$. Then we adopt the attention mechanism to model the cross-entity interaction of these superpixels obtaining $F^{CR}\in \mathbb{R}^{C \times D}$, which could be formulated as:
\begin{equation}
\label{equ-fa}
\begin{aligned}
    F^{CR} &= {\rm MultiHead}(F^{IN}, F^{IN},F^{IN})\\
    &={\rm Concat}({\rm head}_1,\cdots,{\rm head}_h)W^O,
\end{aligned}
\end{equation}
\begin{equation}
{\rm head}_i=      {\rm Attention} (F^{IN} W^Q, F^{IN} W^K, F^{IN} W^V),
\end{equation}
where $W^O \in \mathbb{R}^{D^\prime \times D} $, $W^Q \in \mathbb{R}^{D \times D^\prime} $, $W^K \in \mathbb{R}^{D \times D^\prime} $, and $W^V \in \mathbb{R}^{D \times D^\prime} $ are projection matrices. $D^\prime$ is the feature dimension of the attention module.
In this way, we achieve the interaction modeling among entities because each clustering center corresponds to a specific superpixel. Before the relation prediction module, we inject interaction information earlier at the superpixel level.

Finally, we dispatch the superpixel-level feature $F^{CR}$ into points in the corresponding superpixel to obtain $F^{SIL} \in \mathbb{R}^{M \times D}$ as the output of the SIL block. To be specific, the dispatching process can be formulated as: 
\begin{equation}
\label{equ-fsil}
    F^{SIL}_i = F_i + sim(F_i, F^c_j) \times H^{\prime \prime}(F^{CR}_j),
\end{equation}
where $j=P^{QT}_i$ indicates the index of the cluster to which point $Q_i$ belongs according to the clustering result.

\subsection{Superpixel-Based Interaction Learning Network}
Applying SIL blocks in scene graph generation, we adopt a hierarchical superpixel-based interaction learning network. 
During this hierarchical process, the degree of interaction increases gradually. After the calculation of all SIL blocks, we obtain features with superpixel-level interaction information, which is finer than the existing box-level features. 
We replace $F$ with the output $F^{SIL}$ of the last SIL block as the new feature map of the input image. Features of entities are extracted using $F^{SIL}$ and their bounding boxes.
Then we continue the remaining steps of the traditional SGG methods with these features that contain superpixel-level interaction information.
Finally, the detected entities and the classified relations construct the output scene graph as usual.

\section{Experiments}
We report the experimental results of SIL on multiple scene graph generation baselines in this section.
Firstly, we introduce the dataset preparation and experimental configurations scheme utilized in our study. 
Secondly, the implementation details of SIL in experiments are provided. 
Finally, we provide and analyze the experimental results, which include overall assessments, ablation studies, and the visualization of SIL. 

\begin{table*}[htbp]
\centering
\caption{Comparison with existing state-of-the-art methods on the VG dataset over three tasks.}
\label{tab-sota}
\resizebox{\textwidth}{!}{
\begin{tabular}{c|c|ccc|ccc|ccc}
\hline
\multirow{2}{*}{Model} & \multirow{2}{*}{Type} & \multicolumn{3}{c|}{PredCls} & \multicolumn{3}{c|}{SGCls} & \multicolumn{3}{c}{SGGen} \\
&& mR@20   & mR@50   & mR@100   & mR@20   & mR@50  & mR@100  & mR@20  & mR@50  & mR@100  \\ \hline
VTransE\cite{vtranse}\tiny{\emph{CVPR'17}}  &Box-based        & -  & 14.7& 15.8&    -     & 8.2& 8.7&   -     & 5.0& 6.1    \\
RelDN\cite{reldn}\tiny{\emph{CVPR'19}}  &Box-based      & -   & 15.8    & 17.2     & - & 9.3    & 9.6     & -  & 6.0    & 7.3     \\
GBNet-$\beta$\cite{gbnet}\tiny{\emph{ECCV'20}}  &Box-based     &  -& 22.1    & 24.0    &  - & 12.7   & 13.4    & -  & 7.1    & 8.5     \\
FCSGG\cite{fcsgg}\tiny{\emph{CVPR'21}}  &Box-based        &-    & 6.3     & 7.1      &-   & 3.7    & 4.1     & -& 3.6    & 4.2     \\
IWSL\cite{iwsl}\tiny{\emph{arXiv'22}}&Box-based&      -  & 30.0    & 32.1     & -    & 17.4   & 18.9    & -      & 13.7   & 15.9    \\
IS-GGT\cite{isggt}\tiny{\emph{CVPR'23}}  &Box-based     & - & 26.4    & 31.9     & -  & 15.8   & 18.9    & -      & 9.1    & 11.3    \\
PE-Net\cite{penet}\tiny{\emph{CVPR'23}}  &Box-based     & 25.8 & 31.4 & 33.5 & 15.2   & 18.2   & 19.3    & 9.2    & 12.3   & 14.3    \\
\rowcolor{gray!10} Ours  &Superpixel-based    & \textbf{26.9}    & \textbf{33.1}    &\textbf{35.3}    & \textbf{16.7}   & \textbf{19.9}   & \textbf{20.7}   &\textbf{10.0} &\textbf{13.0} &\textbf{15.1} \\ \hline
\end{tabular}
}
\end{table*}

\subsection{Dataset}
In this paper, experiments were conducted on the Visual Genome (VG) dataset\cite{vg} and the Open Images V6 (OI V6) dataset\cite{oi}, all of which are the benchmark datasets for scene graph generation. Visual Genome contains 108,077 annotated images. Following previous works\cite{bgnn}, we adopt the most frequent 150 entity categories and 50 relation categories in the VG dataset. OI V6 dataset for scene graph generation has 126,368 images for training, 1,813 ones for validation, and 5,322 ones for test. The details of these two datasets are provided in the supplementary material.

\begin{table}[tbp]
\centering
\caption{The performance of methods without or with SIL and comparison with existing state-of-the-art methods on the OI V6 dataset.}
\label{tab-expoi}
 \resizebox{\columnwidth}{!}{
\begin{tabular}{c|l|ll|l}
\hline
Methods  & \multicolumn{1}{c|}{R@50} & \multicolumn{1}{c}{$\rm wmAP_{rel}$} & \multicolumn{1}{c|}{$\rm wmAP_{phr}$} & \multicolumn{1}{c}{$\rm score_{wtd}$} \\ \hline

Motif\cite{motif}\tiny{\emph{CVPR'18}}& 71.6   & 29.9       & 31.6       & 38.9          \\
\rowcolor{gray!10} Motif+SIL   & \textbf{72.3} {\footnotesize \textbf{\textcolor{ACMRed}{(+0.7)}}}  & \textbf{30.4} {\footnotesize \textbf{\textcolor{ACMRed}{(+0.5)}}} & \textbf{31.8} {\footnotesize \textbf{\textcolor{ACMRed}{(+0.2)}}}& \textbf{39.4} {\footnotesize \textbf{\textcolor{ACMRed}{(+0.5)}}}     \\ \hline

G-RCNN\cite{graphrcnn}\tiny{\emph{ECCV'18}} & 74.5       & 33.2       & 34.2       & 41.9      \\
\rowcolor{gray!10} G-RCNN+SIL  & \textbf{75.5} {\footnotesize \textbf{\textcolor{ACMRed}{(+1.0)}}}& \textbf{33.8} {\footnotesize \textbf{\textcolor{ACMRed}{(+0.6)}}}& \textbf{34.7} {\footnotesize \textbf{\textcolor{ACMRed}{(+0.5)}}}& \textbf{42.5} {\footnotesize \textbf{\textcolor{ACMRed}{(+0.6)}}}               \\ \hline

GPS-Net\cite{vctree}\tiny{\emph{CVPR'20}}        & 74.8 & 32.9       & 34.0       & 41.7   \\
\rowcolor{gray!10} GPS-Net+SIL   & \textbf{75.7} {\footnotesize \textbf{\textcolor{ACMRed}{(+0.9)}}}& \textbf{33.9} {\footnotesize \textbf{\textcolor{ACMRed}{(+1.0)}}}& \textbf{34.9} {\footnotesize \textbf{\textcolor{ACMRed}{(+0.9)}}}& \textbf{42.7} {\footnotesize \textbf{\textcolor{ACMRed}{(+1.0)}}}      \\ \hline

BGNN\cite{bgnn}\tiny{\emph{CVPR'21}}  & 75.0          & 33.5       & 34.2       & 42.1       \\
\rowcolor{gray!10} BGNN+SIL      & \textbf{76.4} {\footnotesize \textbf{\textcolor{ACMRed}{(+1.4)}}}& \textbf{34.2} {\footnotesize \textbf{\textcolor{ACMRed}{(+0.7)}}}& \textbf{35.0} {\footnotesize \textbf{\textcolor{ACMRed}{(+0.8)}}}      & \textbf{43.0} {\footnotesize \textbf{\textcolor{ACMRed}{(+0.9)}}}\\ \hline

PE-Net\cite{penet}\tiny{\textit{CVPR'23}} & 76.5&36.6&37.4&44.9 \\
\rowcolor{gray!30} PE-Net+SIL & \textbf{77.1} {\footnotesize \textbf{\textcolor{ACMRed}{(+0.6)}}} & \textbf{37.2} {\footnotesize \textbf{\textcolor{ACMRed}{(+0.6)}}} & \textbf{37.8} {\footnotesize \textbf{\textcolor{ACMRed}{(+0.4)}}} & \textbf{45.5} {\footnotesize \textbf{\textcolor{ACMRed}{(+0.6)}}} \\ \hline
\end{tabular}
}
\vspace{-0.6cm}
\end{table}

\subsection{Evaluation Datasets and Metrics}
\subsubsection{Visual Genome}
In the same way as \cite{bgnn}, we evaluate SGG methods with or without SIL on the VG dataset under three tasks: predicate classification (PredCls), scene graph classification (SGCls), and scene graph detection (SGGen). 
In PredCls, the bounding boxes and object categories are both provided, and the model needs to predict relation categories. 
In the SGCls task, only the bounding boxes of entities are given. 
In the SGGen task, no information about the positions and categories of entities is provided at all. The model relies on itself to detect entities and predict the relations.
We take Mean Recall@K (i.e. mR@K) as the evaluation metric for the scene graph generation task in order to alleviate the misleading effects of Recall on the performance caused by long-tailed datasets.

\subsubsection{Open Images V6}
Following \cite{oi,reldn,bgnn}, we evaluate different methods using $Recall@50$ ($R@50$), weighted mean AP of relations ($wmAP_{rel}$), and weighted mean AP of phrases ($wmAP_{phr}$). In line with the standard evaluation of Open Images, we report the weight metric $score_{wtd}$ which is obtained as $0.2 \times R@50 + 0.4 \times wmAP_{rel} + 0.4 \times wmAP_{phr}$ following \cite{oi,reldn}.

\subsection{Implementation Details}
To ensure the fairness of comparison, we employ the same visual backbone as the previous methods, that is, we use ResNeXt-101-FPN \cite{resnet} as the backbone network and FasterRCNN \cite{fasterrcnn} as the object detector. We keep the model parameters of these visual backbone modules frozen in the training process of SGG models. In this way, we ensure that the improvement in experimental results comes from the conversion and enhancement of the same basic visual features, rather than using more favorable ones. 
As for our SIL block, we adopt the multi-head design \cite{attention} to enhance the context clustering step following \cite{coc}. The embedding dimension is 256 during the superpixel clustering. Before the intra-entity interaction module, we perform feature dimension transformation on the output of superpixel clustering using a simple Multilayer Perceptron (MLP) network. The feature dimension used in the intermediate layer of the MLP is 1024. We use the corresponding feature in the output of ResNeXt-101-FPN backbone as the reference visual feature $g_k$ we used in Eq.~\eqref{equ-fagg}. In the cross-entity interaction module, we use a 2-layer transformer encoder. The encoder is multi-head style and each head is expressed with the embedding dimension of 32. The dropout is 0.1 in the encoder. 
We keep the optimizer, the scheduler, and the number of iterations the same with each baseline. Experiments in this paper are implemented with PyTorch. We use an NVIDIA GeForce RTX 3090 GPU to train the SGG models.

\begin{table}[tbp]
\centering
\caption{Ablation study on the component of our SIL block. SC denotes the Superpixel Clustering step. IEI denotes the Intra-Entity Interaction module. $\CIRCLE$ means using weighted superpixels in IEI. $\Circle$ means IEI without the superpixel weights. CEI denotes the Cross-Entity Interaction module.}
\label{tab-ablation}
\begin{tabular}{ccc|lll}
\hline
SC & IEI & CEI & \multicolumn{1}{c}{mR@20} & \multicolumn{1}{c}{mR@50} & \multicolumn{1}{c}{mR@100} \\ \hline
- & - & - & 12.31  & 15.03 & 16.10 \\
\Checkmark & - & - & 12.59 {\footnotesize \textbf{\textcolor{ACMRed}{(+0.28)}}} & 15.54 {\footnotesize \textbf{\textcolor{ACMRed}{(+0.51)}}} & 16.75 {\footnotesize \textbf{\textcolor{ACMRed}{(+0.65)}}} \\
\Checkmark  & $\CIRCLE$ & - & 12.61 {\footnotesize \textbf{\textcolor{ACMRed}{(+0.30)}}} & 15.76 {\footnotesize \textbf{\textcolor{ACMRed}{(+0.73)}}} & 17.08 {\footnotesize \textbf{\textcolor{ACMRed}{(+0.98)}}} \\
\Checkmark & - & \Checkmark & 13.16 {\footnotesize \textbf{\textcolor{ACMRed}{(+0.85)}}} & 16.40 {\footnotesize \textbf{\textcolor{ACMRed}{(+1.37)}}} & 17.70 {\footnotesize \textbf{\textcolor{ACMRed}{(+1.60)}}} \\
\Checkmark & $\Circle$ & \Checkmark & 13.12 {\footnotesize \textbf{\textcolor{ACMRed}{(+0.81)}}} & 16.52 {\footnotesize \textbf{\textcolor{ACMRed}{(+1.49)}}} & 17.78 {\footnotesize \textbf{\textcolor{ACMRed}{(+1.68)}}} \\
\rowcolor{gray!10} \Checkmark & $\CIRCLE$ & \Checkmark & 13.62 {\footnotesize \textbf{\textcolor{ACMRed}{(+1.31)}}} & 16.91 {\footnotesize \textbf{\textcolor{ACMRed}{(+1.88)}}} & 18.36 {\footnotesize \textbf{\textcolor{ACMRed}{(+2.26)}}} \\ \hline
\end{tabular}%
\end{table}

\begin{figure}[tbp]
    \centering
    \includegraphics[width=7cm]{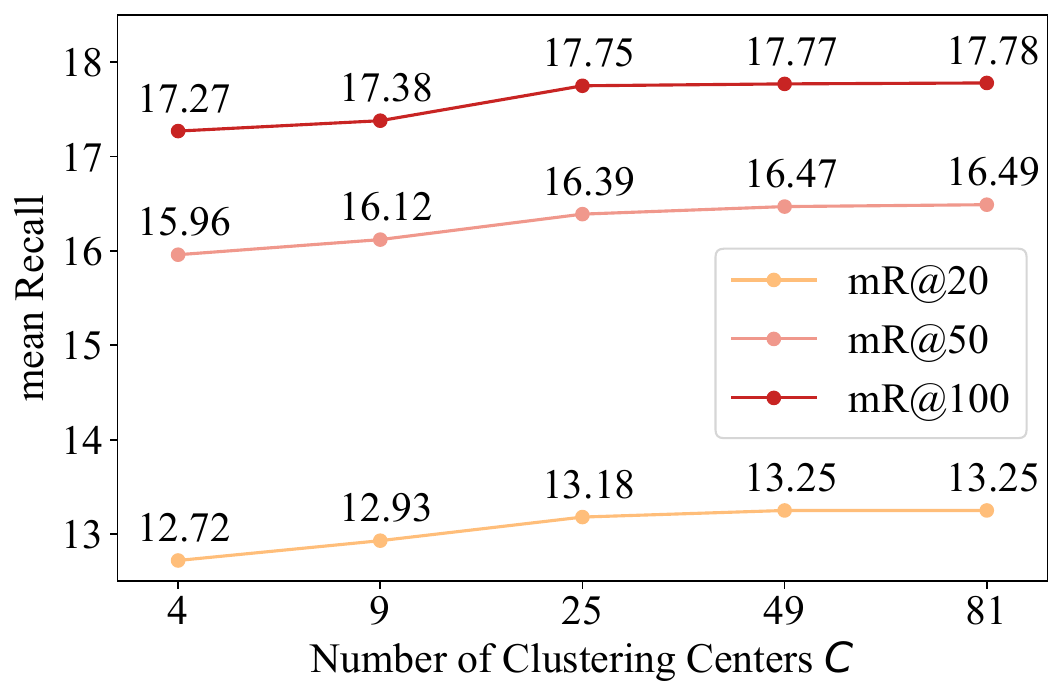}
    \vspace{-0.3cm}
    \caption{The influence of the number of clustering centers $C$ in the SIL block.}
    \label{fig-ncc}
    \vspace{-0.3cm}
\end{figure}

\subsection{Overall Performance Assessments}
\subsubsection{Visual Genome}
\label{4.4.1}
To demonstrate the effectiveness and generalization ability of our SIL block, we conducted experiments on various scene graph generation methods. The results on the VG dataset are shown in Table \ref{tab-exp}. In such the baselines, Motif\cite{motif}, G-RCNN\cite{graphrcnn}, KERN\cite{kern}, and VCTree\cite{vctree} are classical methods for SGG. GPS-Net\cite{gpsnet}, BGNN\cite{bgnn}, and PENet\cite{penet} are relatively lately-published methods. These baselines contain methods with various implementations including Recurrent Neural Networks (RNN), Graph Neural Networks (GNN), and Graph Convolutional Networks (GCN). For the experiments of original baseline models, we report the results according to \cite{unbiased,bgnn}.
As we can see, the SIL block enables all these scene graph generation methods to achieve very stable improvements. No matter what structure the baseline is based on, we could use the SIL block to convert the original box-level features to superpixel-level ones, which could model the fine-grained interactions among entities. As shown in Table \ref{tab-exp}, on the PredCls task which attaches great importance to the relation prediction, the SIL block has brought an average improvement of 2.0\% to the mean Recall metric of these baselines. Without any additional sources of information and supervision signals, SIL improves these baselines with the help of newly introduced superpixel-level feature extraction and fine-grained interaction.

\subsubsection{Open Images V6}
Similarly to Sec. \ref{4.4.1}, we perform experiments on the OI V6 dataset with Motif\cite{motif}, G-RCNN\cite{graphrcnn}, GPS-Net\cite{gpsnet}, BGNN\cite{bgnn}, and PE-Net\cite{penet} as baselines. The experimental results are shown in Table \ref{tab-expoi}. The results of baselines in Table \ref{tab-expoi} are borrowed from \cite{bgnn,penet}. Baselines are improved stably thanks to the superpixel-level features and interactions our SIL block achieved. The complete experimental results table is provided in the supplementary material.

\subsubsection{Comparison with the State of the Art}
We validate the effectiveness of our SIL block by comparison with existing state-of-the-art (SOTA) methods.
The comparison results with SOTA on the VG and OI V6 datasets are shown in Table \ref{tab-sota} and Table \ref{tab-expoi} respectively. 
In Table \ref{tab-sota}, Results of IS-GGT\cite{isggt} and IWSL\cite{iwsl} are borrowed from corresponding papers. We report the results of PE-net\cite{penet} according to the model provided by the author. As we can see, our model with the SIL block outperforms state-of-the-art methods on the VG and OI V6 datasets. 

\subsection{Ablation Study}
In order to confirm the effect of each component in our SIL block, we perform ablation experiments. Experiments of our ablation studies are all done on the PredCls task on the Visual Genome dataset. We choose the typical Motif\cite{motif} model as the baseline to facilitate the observation of the effects of different components. 
\begin{figure}[tbp]
    \centering
    \includegraphics[width=7cm]{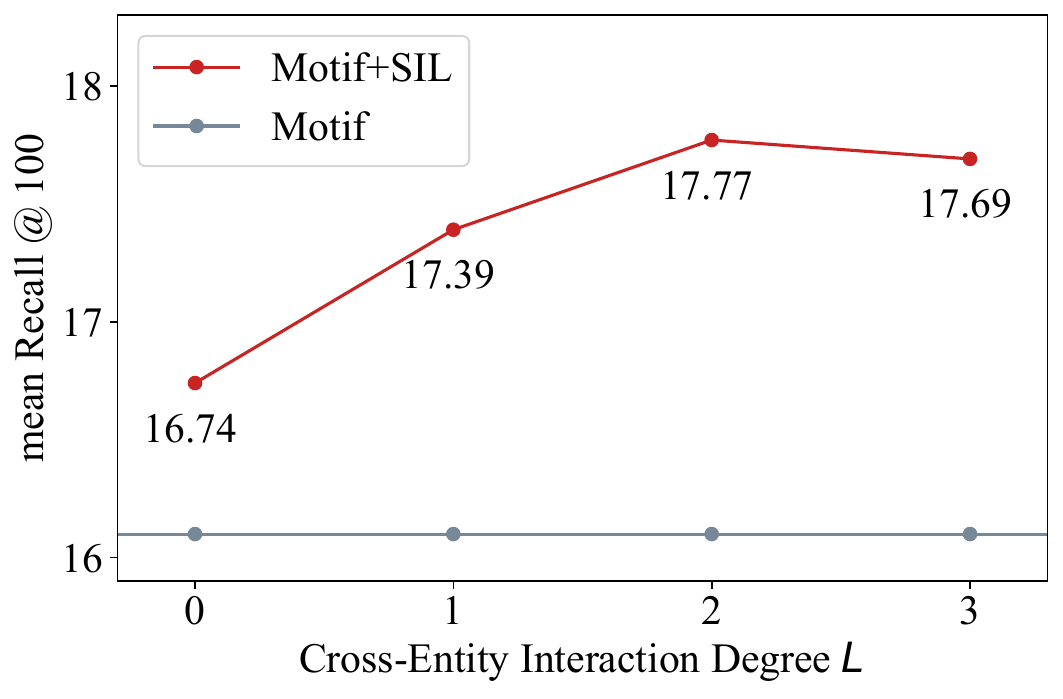}
    \vspace{-0.3cm}
    \caption{The influence of varying interaction degree $L$ in the SIL block.}
    \label{fig-transformer}
    \vspace{-0.5cm}
\end{figure}
\subsubsection{Component Ablation}
In order to illustrate the effect of each component of our SIL block, we conduct ablation experiments on the three main components of the SIL block, namely Superpixel Clustering, Intra-Entity Interaction, and Cross-Entity Interaction. The ablation study results are shown in Table \ref{tab-ablation}. As we can see, each module contributes to the improvement of the Motif baseline. 

In Table \ref{tab-ablation}, the first row denotes the results of Motif we reproduced according to \cite{motif,unbiased}. 
In the second row with only Superpixel Clustering, we extract visual features in the form of superpixel, without any explicit superpixel-level interaction operations. In this case, we still achieve improvements over the original Motif model, which confirms the benefits of superpixel-level features for SGG.
The third and fourth rows demonstrate the necessity of Intra-Entity Interaction and Cross-Entity Interaction respectively.
Finally, with the complete SIL block, we achieve the result in the last row of Table \ref{tab-ablation}. 
Besides, we also evaluate the SIL block without superpixel weights, that is, assign the same weights to superpixels in Eq.~\eqref{equ-weight}. The corresponding experimental results are shown in the fifth row of Table \ref{tab-ablation}. The comparison between the fifth row and the last row demonstrates that our differentiation of superpixels inside each entity helps SGG indeed.

\subsubsection{The Number of Clustering Centers.}
In Fig. \ref{fig-block}(a), the input is viewed as the sets of points. We perform clustering on the points to obtain superpixels. In the process, we propose $C$ centers denoted by $T$ as the clustering centers as mentioned in \ref{sec-3.1}. 
The number of clustering centers directly determines the number of superpixels into which the input is divided. In order to investigate the influence of the number of clustering centers, we have trials of SIL block with 4, 9, 25, 49, and 81 centers, respectively. Considering memory constraints, we adopt 256-D MLP and 4 heads in Superpixel Clustering in this ablation experiment. The results are shown in Fig. \ref{fig-ncc} where $C$ denotes the number of centers we pre-defined. According to Fig. \ref{fig-ncc}, as the number of clustering centers increased from 4 to 9, 25, and 49, the performance of the models improves gradually and slightly. SIL is not highly sensitive to the number of clustering centers. Therefore, through making a trade-off between model performance and program memory, we chose 49 as the vanilla number of clustering centers in this paper.

\subsubsection{The Influence of Cross-entity Interaction Degree} 
In the SIL block, we perform the cross-entity interaction as shown in Fig. \ref{fig-block}(c) employing a Transformer encoder with the attention mechanism. The more layer we use in the encoder, the more interaction information entities obtain. In order to demonstrate the effect of our superpixel-level interaction learning, we perform experiments with different degrees of interaction denoted by $L$. 
The results are given in Fig. \ref{fig-transformer}. We use 256-D MLP and 4 heads in Superpixel Clustering here considering memory constraints. Larger values on the horizontal axis mean more encoder layers and deeper superpixel-level interaction. There is no superpixel-level interaction modeling in the baseline Motif, therefore the performance of the baseline doesn't vary with the increasing degree of interaction. The 0-layer result is obtained without cross-entity interaction. As the interaction degree increases from 0 to 2, the performance improves gradually. However, too frequent interactions would lead to information confusion as shown in the results of a 3-layer encoder in SIL.

\begin{figure}[tbp]
    \centering
    \includegraphics[width=\linewidth]{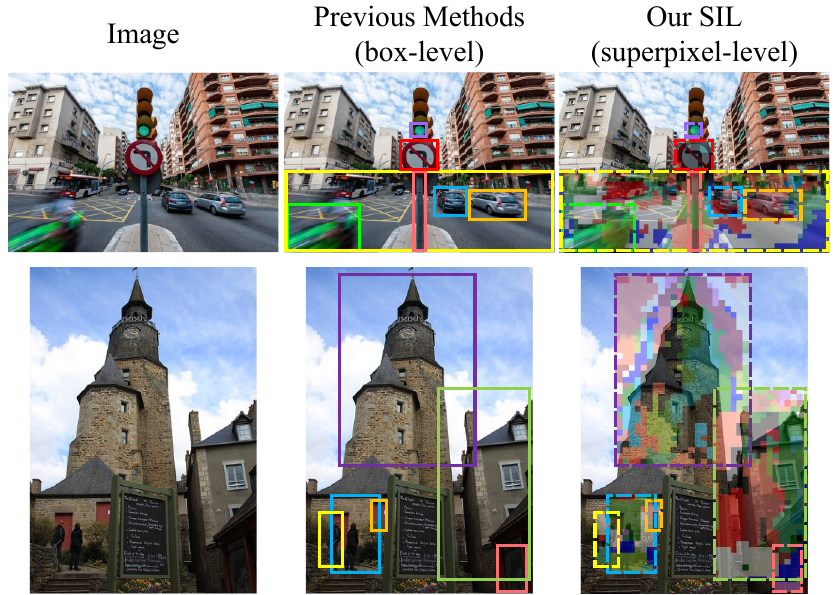}
    \caption{Visualization of clustering results. The first column shows the original image. The second column shows the bounding boxes. The last column shows the superpixels inside bounding boxes. Masks of different colors denote different clusters.}
    \label{fig-visCluster}
\vspace{-0.3cm}
\end{figure}

\subsection{Visualization}
\subsubsection{Clustering Result}
Clustering is a crucial algorithm used in the SIL block. We cluster the points in the input image or feature map to obtain superpixels as the expression for SGG. We visualize the clustering results in Fig. \ref{fig-visCluster}. 
The second column displays the rough boxes used in existing methods, while the last column shows the fine-grained superpixels used in our SIL block. Such specific superpixel-level information endows our approach with differentiated visual features inside each entity and the base for fine-grained interaction, and therefore helps SGG. As we can see from the last column, the Superpixel Clustering module can effectively aggregate similar pixels. The superpixels generated by clustering imply the boundaries of entities clearly. For example, the outline of the tower in the second row is delineated by clustering. 

\begin{figure}[tbp]
    \centering
    \includegraphics[width=\linewidth]{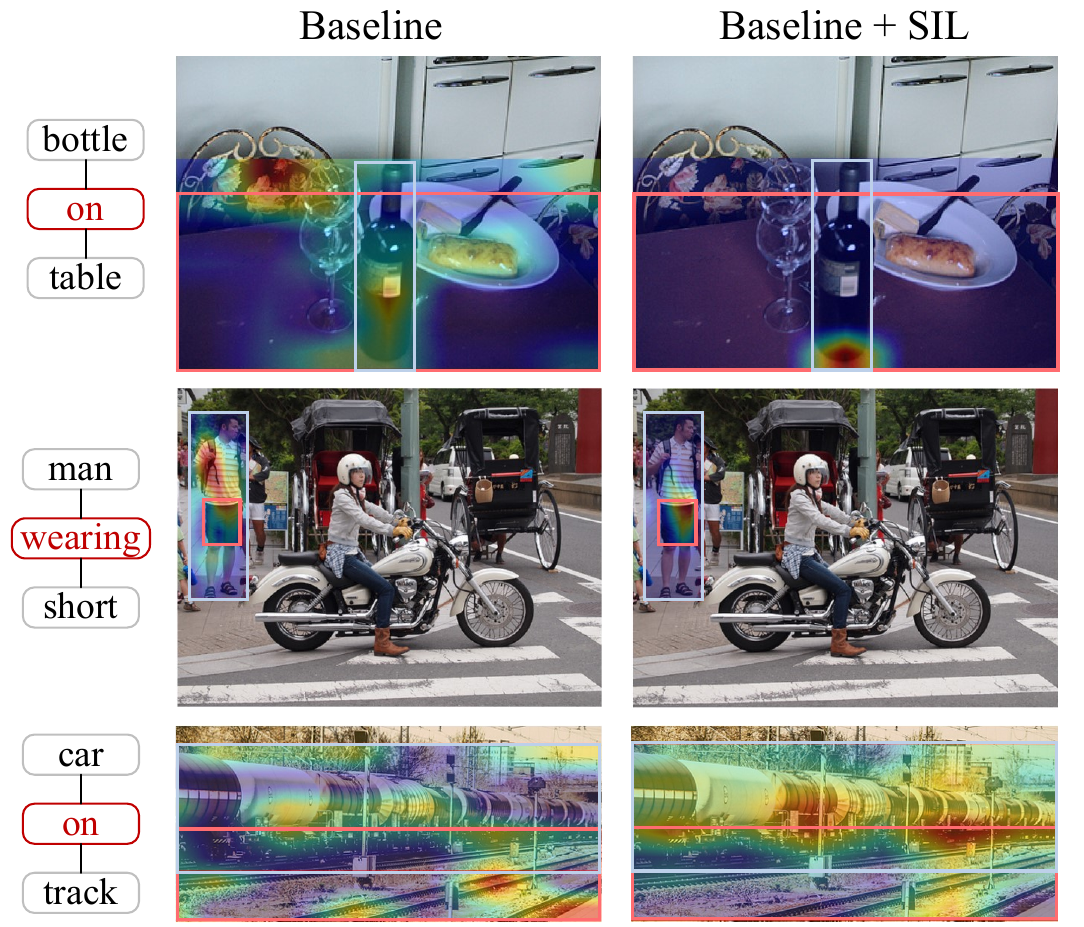}
    \caption{Comparisons of GradCAM on features without/with the SIL block. Red words indicate the relation to which GradCAM is applied.}
    \label{fig-visGradCAM}
\end{figure}

\subsubsection{Relation Activation}
We perform GradCAM \cite{gradcam} on the union regions of subjects and objects with the relation in a triplet as the goal. We adopt the Motif model as the baseline. The GradCAM results are shown in Fig. \ref{fig-visGradCAM}. With the clustering algorithm, our method extracts superpixel-level features and interactions. Therefore, our method could capture the specific interactive sub-regions corresponding to the relations for scene graph generation. For example, in the third case of Fig. \ref{fig-visGradCAM}, the baseline is confused by the ambiguous information inside boxes, while our method captures the interaction position of the car on the track accurately.

\section{Conclusion}
In this paper, we propose the Superpixel-based Interaction Learning (SIL) block to convert the traditional box-based pipeline for scene graph generation into a superpixel-based one. In SIL, any input image is regarded as a set of points. We use the clustering algorithm to aggregate all the points into superpixels and extract deep features of them. Furthermore, we perform intra-entity interaction to ensure the differentiation of superpixels in an entity. We design a cross-entity interaction module, so as to get more fine-grained interaction information among entities for the generation of scene graphs. We evaluate SIL on various scene graph generation baselines. The experimental results demonstrate the effectiveness and generalization ability of SIL on the scene graph generation task. With experimental validation, our SIL outperforms previous state-of-the-art methods on SGG.

\begin{acks}
This work was supported in part by the National Science Foundation of China (NSFC) under Grant No. 62176134, by a grant from the Institute Guo Qiang (2019GQG0002), Tsinghua University, and by research and application on AI technologies for smart mobility funded by SAIC Motor.

\end{acks}

\clearpage
\bibliographystyle{ACM-Reference-Format}
\bibliography{root}

\clearpage
\appendix
\section{Datasets and Evaluation Metrics}

\subsection{Datasets}
\paragraph{Visual Genome}
Visual Genome(VG) is the benchmark dataset of Scene Graph Generation (SGG). On average, there are 38 entities and 22 relation instances in each image.
The dataset is partitioned into a training set containing 70\% of the images, and a testing set containing the remaining 30\% images. 5k images from the training set are used for validation. 
\paragraph{Open Images V6}
The Open Image V6 (OI V6) dataset is also used for the evaluation of scene graph generation methods. On average, each image in OI V6 contains 4.1 entities and 2.8 relation instances.
There are 301 entity categories and 31 relation categories in total.

\subsection{Evaluation Metrics}
\paragraph{Visual Genome}
In line with \cite{bgnn}, we take Mean Recall@$K$ (i.e. $mR@K$) as evaluation metrics for the scene graph generation task. For each image, the ground truth triplets are matched with the top-$K$ predicted triplets. The recall rate is calculated as the ratio of the number of matched ground truth triplets to the total number of ground truth triplets in the image. Finally, we obtain mean Recall@$K$ by averaging the recall rate over all types of predicates.
\paragraph{Open Images V6}
Following the settings in \cite{oi,reldn,bgnn}, we evaluate different methods using $Recall@50$ ($R@50$), weighted mean AP of relations ($wmAP_{rel}$), and weighted mean AP of phrases ($wmAP_{phr}$). $R@50$ has the same definition as that used in VG evaluation. As for <subject-predicate-object> triplets, $AP_{rel}$ assesses the subject box and the object box, along with three labels of the subject, the object, and the predicate in a triplet. Compared to $AP_{rel}$, $AP_{phr}$ evaluates the union bounding box of the subject box and the object box, and the same three labels. We report the weighted average of $AP_{rel}$ and $AP_{phr}$ of each relation category as $wmAP_{rel}$ and $wmAP_{phr}$ to mitigate the impact of the dataset bias on evaluation. 

\section{Implementation Details}
\subsection{CodeBase}
We implemented the proposed Superpixel-based Interaction Learning (SIL) and conducted comparative experiments on the baselines using the general codebase for SGG\footnote{https://github.com/KaihuaTang/Scene-Graph-Benchmark.pytorch\label{codebase}}, the official codebase of BGNN\footnote{https://github.com/SHTUPLUS/PySGG} and PE-Net\footnote{https://github.com/VL-Group/PENET}.

\subsection{Training Details}
We use the commonly used pretrained object detector\footref{codebase} which is provided by the author of \cite{unbiased}. The model parameters of these visual backbone modules are frozen in the training process of SGG models. We did not modify the relation prediction modules in the baseline methods.

\section{Results on Open Images V6}
We presented the performance of methods without or with SIL on the OI V6 dataset in the main paper. The complete experimental results are shown in Table \ref{supp-expoi}.

\begin{table}[htbp]
\centering
\caption{The comparison results of methods without or with SIL on the OI V6 dataset.}
\label{supp-expoi}
 \resizebox{\columnwidth}{!}{
\begin{tabular}{c|l|ll|l}
\hline
Methods  & \multicolumn{1}{c|}{R@50} & \multicolumn{1}{c}{$\rm wmAP_{rel}$} & \multicolumn{1}{c|}{$\rm wmAP_{phr}$} & \multicolumn{1}{c}{$\rm score_{wtd}$} \\ \hline
RelDN\cite{reldn}\tiny{\textit{CVPR'19}} & 73.1 & 32.2 & 33.4 & 40.8 \\
VCTree\cite{vctree}\tiny{\textit{CVPR'19}} & 74.1 & 34.2 & 33.1 & 40.2 \\
IWSL\cite{iwsl}\tiny{\textit{arXiv'22}} & 74.7 & 33.1 & 34.3 & 41.9 \\ \hline

Motif\cite{motif}\tiny{\emph{CVPR'18}}& 71.6   & 29.9       & 31.6       & 38.9          \\
\rowcolor{gray!10} Motif+SIL   & \textbf{72.3} {\footnotesize \textbf{\textcolor{ACMRed}{(+0.7)}}}  & \textbf{30.4} {\footnotesize \textbf{\textcolor{ACMRed}{(+0.5)}}} & \textbf{31.8} {\footnotesize \textbf{\textcolor{ACMRed}{(+0.2)}}}& \textbf{39.4} {\footnotesize \textbf{\textcolor{ACMRed}{(+0.5)}}}     \\ \hline

G-RCNN\cite{graphrcnn}\tiny{\emph{ECCV'18}} & 74.5       & 33.2       & 34.2       & 41.9      \\
\rowcolor{gray!10} G-RCNN+SIL  & \textbf{75.5} {\footnotesize \textbf{\textcolor{ACMRed}{(+1.0)}}}& \textbf{33.8} {\footnotesize \textbf{\textcolor{ACMRed}{(+0.6)}}}& \textbf{34.7} {\footnotesize \textbf{\textcolor{ACMRed}{(+0.5)}}}& \textbf{42.5} {\footnotesize \textbf{\textcolor{ACMRed}{(+0.6)}}}               \\ \hline

GPS-Net\cite{vctree}\tiny{\emph{CVPR'20}}        & 74.8 & 32.9       & 34.0       & 41.7   \\
\rowcolor{gray!10} GPS-Net+SIL   & \textbf{75.7} {\footnotesize \textbf{\textcolor{ACMRed}{(+0.9)}}}& \textbf{33.9} {\footnotesize \textbf{\textcolor{ACMRed}{(+1.0)}}}& \textbf{34.9} {\footnotesize \textbf{\textcolor{ACMRed}{(+0.9)}}}& \textbf{42.7} {\footnotesize \textbf{\textcolor{ACMRed}{(+1.0)}}}      \\ \hline

BGNN\cite{bgnn}\tiny{\emph{CVPR'21}}  & 75.0          & 33.5       & 34.2       & 42.1       \\
\rowcolor{gray!10} BGNN+SIL      & \textbf{76.4} {\footnotesize \textbf{\textcolor{ACMRed}{(+1.4)}}}& \textbf{34.2} {\footnotesize \textbf{\textcolor{ACMRed}{(+0.7)}}}& \textbf{35.0} {\footnotesize \textbf{\textcolor{ACMRed}{(+0.8)}}}      & \textbf{43.0} {\footnotesize \textbf{\textcolor{ACMRed}{(+0.9)}}}\\ \hline

PE-Net\cite{penet}\tiny{\textit{CVPR'23}} & 76.5&36.6&37.4&44.9 \\
\rowcolor{gray!30} PE-Net+SIL & \textbf{77.1} {\footnotesize \textbf{\textcolor{ACMRed}{(+0.6)}}} & \textbf{37.2} {\footnotesize \textbf{\textcolor{ACMRed}{(+0.6)}}} & \textbf{37.8} {\footnotesize \textbf{\textcolor{ACMRed}{(+0.4)}}} & \textbf{45.5} {\footnotesize \textbf{\textcolor{ACMRed}{(+0.6)}}} \\ \hline
\end{tabular}
}
\end{table}
\section{Visualization}
\begin{figure*}[htbp]
    \centering
    \includegraphics[width=17cm]{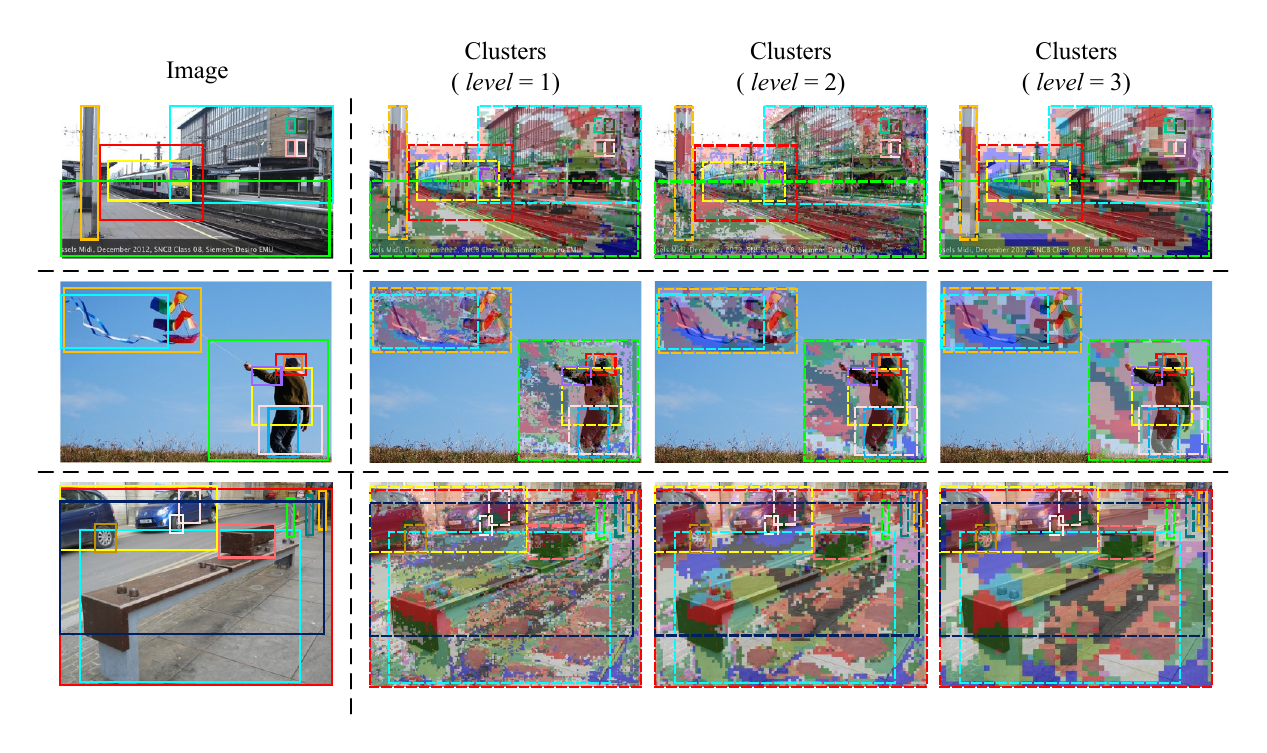}
    \vspace{-0.4cm}
    \caption{Visualization of the clustering results in SIL. The first column shows the original image and boxes of entities in each image. The other columns show the clustering results of visual features at different levels.}
    \label{supp-visCluster}
\end{figure*}
\vspace{0.5cm}
\subsection{Clustering Result}
In the Superpixel Clustering module of our SIL block, we cluster the points in the input image or feature map to obtain superpixels as the expression for SGG. We visualize the clustering results of Motif\cite{motif} equipped with SIL on the PredCls task in Fig. \ref{supp-visCluster}. As we can see, the Superpixel Clustering module could aggregate similar pixels effectively. As the clustering gets deeper, the clustered superpixels become more holistic and imply larger sub-regions in the original image. 
The superpixels generated by clustering could represent the boundary of entities clearly and enable the modeling of fine-grained interactions.

\begin{figure*}[htbp]
    \centering
    \includegraphics[width=15cm]{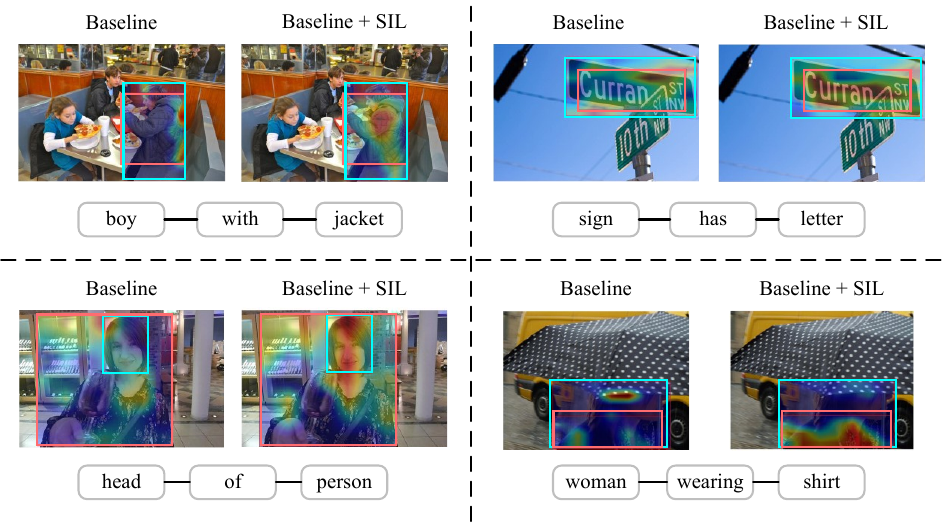}
    \caption{More comparisons of GradCAM on features without/with the SIL block. Blue boxes indicate subjects in triplets. Pink boxes indicate objects.}
    \label{supp-visGradCAM}
\end{figure*}

\subsection{Relation Activation}
We perform GradCAM visualization to better understand the effect of our modeling of fine-grained interactions. Specifically, we perform GradCAM \cite{gradcam} on the union regions of subjects and objects with the predicate in a triplet as the goal. We adopt the Motif model as the baseline for the PredCls task. 
The GradCAM results are shown in Fig. \ref{supp-visGradCAM}. We compare the GradCAM results of the baseline without and with SIL, and annotate the corresponding triplets.
The baseline method with coarse-grained interactions is inaccurate in capturing the specific location where interactions occur.
With the fine-grained interaction modeling at the superpixel level in the proposed SIL, our method could capture the specific interactive sub-regions corresponding to the relations for scene graph generation. 

\section{Broader Impact and Limitations} 
We propose SIL converting the previous box-level SGG pipeline into a superpixel-based one with fine-grained interactions. 
The presented SIL method can be integrated into any existing box-level model in a plug-and-play way. However, while improving the performance on scene graph generation, frequent fine-grained interactions lead to more time consumption. 
Therefore, a more efficient and sparse fine-grained interaction approach should be explored in further work.

\end{document}